\documentclass[lettersize,journal]{IEEEtran}
\usepackage{amsmath,amsfonts}
\usepackage{algorithmicx}    
\usepackage{algorithm}
\usepackage{setspace}
\usepackage{algpseudocode}
\usepackage{array}
\usepackage[caption=false,font=normalsize,labelfont=sf,textfont=sf]{subfig}
\usepackage{textcomp}
\usepackage{stfloats}
\usepackage{url}
\usepackage{verbatim}
\usepackage{graphicx}
\usepackage{cite}
\usepackage{epsfig}
\usepackage{graphicx}
\usepackage{xcolor}
\usepackage{caption}
\usepackage{soul}
\usepackage{tabularx}

\newcommand{\change}[1]{{\textcolor{black}{#1}}}
\newcommand{\newchange}[1]{{\textcolor{black}{#1}}}

\begin{document}

\title{Leveraging Haptic Feedback to Improve \\Data Quality and Quantity for \\Deep Imitation Learning Models}

\author{Catie Cuan$^{1}$, Allison Okamura$^{1}$, and Mohi Khansari$^{2}$
\thanks{*This work was supported by Everyday Robots and Google.}
\thanks{$^{1}$Catie Cuan and Allison Okamura are with the Department of Mechanical Engineering, Stanford University, Stanford, CA 94305, USA
        {\tt\small ccuan@stanford.edu, aokamura@stanford.edu}}%
\thanks{$^{2}$Mohi Khansari is with Cruise, San Francisco, CA 94107, USA, worked completed while at Everyday Robots
        {\tt\small mohi.khansari@gmail.com}}%
}

\maketitle

\begin{abstract}
Learning from demonstration is a proven technique to teach robots new skills. Data quality and quantity play a critical role in \change{the performance of models trained using data collected from human demonstrations}. In this paper we \change{enhance} an existing teleoperation data collection system with real-time haptic feedback \change{to the human demonstrators}; we observe improvements in the collected data throughput and \change{in the performance of autonomous policies using models trained with the data}. Our experimental testbed was a mobile manipulator robot that opened doors with latch handles. Evaluation of teleoperated data collection on eight real conference room doors found that adding haptic feedback improved data throughput by 6\%. We additionally used the collected data to train six image-based deep imitation learning models, three with haptic feedback and three without it. These models were used to implement autonomous door-opening with the same type of robot used during data collection. A policy from a imitation learning model trained with data \change{collected while the human demonstrators received haptic feedback} performed on average 11\% better than its counterpart trained with \change{data collected without haptic feedback}, indicating that haptic feedback \change{provided during data collection resulted in improved autonomous policies}.
\end{abstract}

\vspace{-5mm}

\section{INTRODUCTION}

Learning from demonstration \change{(LfD)} is a promising approach to teach robots new skills based on a set of expert demonstrations \cite{argall2009survey, fang2019survey, schaal1999imitation}. In this paradigm, the learned policy benefits from the demonstrators' prior knowledge about a task (in contrast to self exploration) in order to learn the task with a reasonable number of demonstrations (preferably not more than a couple of hours). Demonstrators can be humans, other robots, or other simulated agents, and demonstrations can be collected in a variety of modes, including teleoperation, third-person video recording, and kinesthetic interactions. To improve the ease of filtering the data, demonstrators may automatically tag the data they are collecting as success or failure. Once this data is collected, various techniques such as behavior cloning \cite{torabi2018behavioral} are used to train a model that is then deployed as a policy by an autonomous agent.

\begin{figure*}[h!]
\centering
\includegraphics[width=\linewidth]{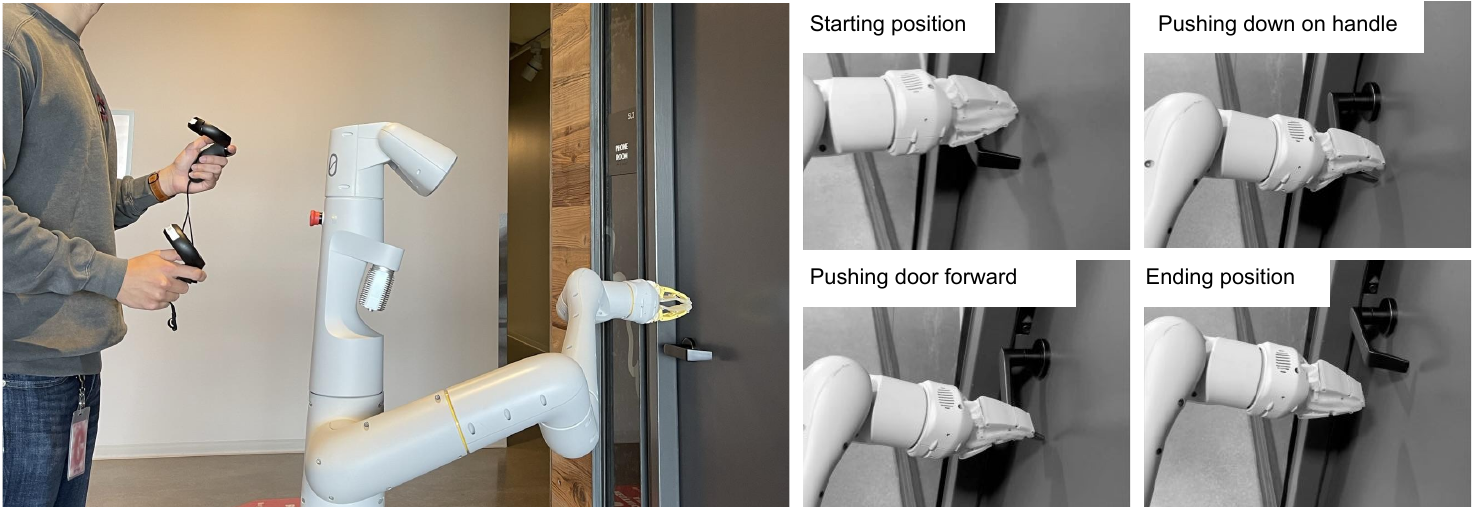}
\caption{\change{We measured how haptic feedback altered task performance in terms of data throughput and data quality in a latch door opening task, and evaluated the effect of the resultant datasets on the autonomous performance of learned policies. Left: A demonstrator teleoperated a prototype mobile manipulator robot. Right: Close-up progression.}}
\vspace{-5mm}
\label{push_door_setup}
\end{figure*}

In visual imitation learning for robotic manipulation, images (observations) and robot movements (actions) gathered from expert demonstrations are separated into observation and action pairs. The demonstrations are captured in the variable $\tau^* = (s_0, a_0, s_1, a_1, ..., s_{T-1}, a_{T-1}, s_T)$. The learned policy $\pi(a\vert s)$ outputs a continuous action $a \in A$ given an image $s \in S$. The aim is to learn a policy $\pi(a\vert s)$ that imitates the expert policy $\pi^*(a|s)$ represented in the demonstrations. Towards this end, teleoperation is an effective mode of data collection because (1) egocentric cameras on the robot can see the task (in contrast to direct human demonstration, where the demonstrator might physically obstruct the robot's view), and (2) the data is directly representative of the robot's action space and does not need to address the correspondence problem (the correspondence problem is the difficulty in matching different training data representations to the robot's possible action space). There may, however, be a kinematic and force feedback mismatch between the human and the teleoperated robot because the robot's sensors may record contact forces that are not felt by the teleoperator. These mismatching feedback channels between the human and the robot may be mitigated by giving the human demonstrator more real-time feedback about the robot's overall state, thus altering the demonstrator's behavior. Haptic feedback is a promising additional feedback channel for teleoperating physical manipulation tasks and may alter demonstrator behavior, such as the time to collect useful data and the quality or consistency of the data. This alteration in demonstrator behavior can impact the learned policy.

The quality of the collected data plays an important role in generating autonomous actions via behavior cloning techniques \cite{hussein2017imitation}. Imagine two scenarios: 
Scenario A: X hours of high-quality data is collected (in this case, data collected with haptics).
Scenario B: 2X \newchange{hours} of confusing, messy data is collected.
Despite Scenario B having more data, it yields poorer performance because the imprecise data reduces training clarity. Thus, to improve the LfD process and understand the role of added haptic feedback, we address two primary research questions:
\vspace{-1mm}
\begin{enumerate}
    \item Does haptic feedback improve the quality and quantity of data collected during teleoperation \newchange{of a real-world robot}?
    \item Do models trained with data collected with haptic feedback \newchange{of a real-world robot} lead to improved autonomous robot task performance (i.e. resultant policies)?
\end{enumerate} 

To address these research questions, we performed a \change{two-phase study, each targeting one of the questions above, on a physical interaction task of latch door opening (Figure \ref{push_door_setup}). The task is performed via teleoperated human demonstration with and without haptic feedback.}
The results indicated that the use of haptic feedback helped with both the data quality and quantity. It is widely accepted in imitation learning that one can improve robot performance through either better model architecture or better data. In our work, we studied the latter and evaluated it at two levels: (1) Data collection and (2) Learned policy performance. Our contribution comes from not only making the data collection faster/higher yield but also demonstrating that this change in demonstrator behavior manifests in a change in learned policy performance. Our results showed that haptic feedback is useful for robot learning not necessarily because the task required it; but because the quality of the collected data is higher. \change{We demonstrated this benefit at several points in the learning process in order to explore both research questions thoroughly. To our knowledge, this is the first work that studied the effect of haptic feedback during teleoperator example collection in deep imitation learning applications \newchange{in the real world}, from example collection to autonomous policy performance.}
\vspace{-3mm}

\section{Prior Work}

We considered tasks that required physical interaction between the robot end-effector and its environment. This interaction constrains the movement of the robot and results in forces that can be used to generate haptic feedback. Prior work performed deep imitation learning with off-the-shelf virtual reality headsets, leading to high success rates for manipulation tasks like grasping or relocating an object \cite{kim2022memory, seita2020deep, xie2020deep, zhang2018deep}. Related work shows that it is possible to train complex manipulation behaviors from images alone \cite{rahmatizadeh2018vision} and in some cases without explicit examples of the task \cite{jang2022bc}. Certain robotic teleoperation tasks have been performed with only visual feedback, such as steering a robotic vehicle \cite{recchiuto2016visual} or using a wide, omni-viewing camera to move an occluded object \cite{nicolis2018occlusion}. Prior work demonstrated that adding haptic feedback to a visual teleoperation system improved abstract motor skills \cite{morris2007haptic}, and when provided early in training, could enhance performance with surgical simulators \cite{strom2006early}. Depending on the task context, low-frequency haptic feedback  may have led to improved task performance \cite{wildenbeest2012impact}. In other circumstances, cutaneous haptic feedback increased task success \cite{pacchierotti2015cutaneous}. \newchange{One prior work showed the utility of force information in self-supervised context \cite{lee2020making}, and adding force information into an imitation learning model \cite{misimi2018robotic, edmonds2017feeling}.} This paper builds upon these prior works and examines the research question of whether haptic feedback can improve demonstrator performance as well as policy performance in an LfD paradigm.

\begin{figure*}[t]
\vspace{1.5mm}
\centering
\includegraphics[width=\linewidth]{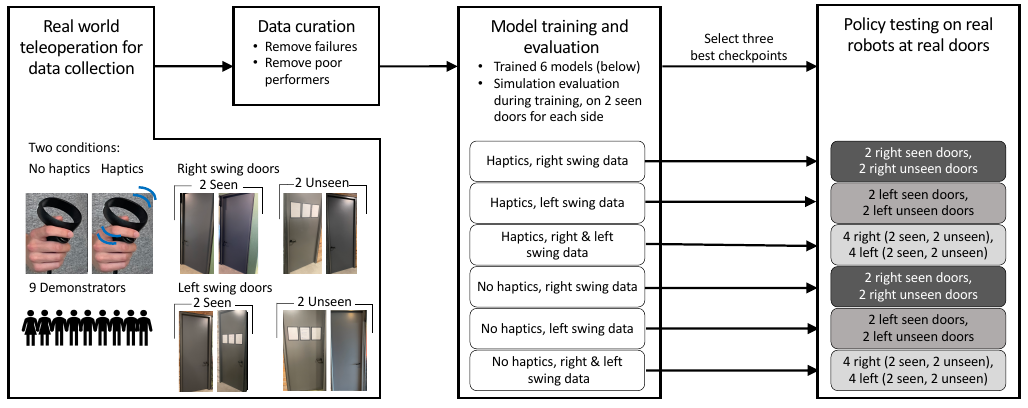}
\caption{Our complete workflow including data collection, curation, training, and policy testing. This paper contributes to the first, second, and fourth steps by introducing haptic feedback to the teleoperator, curating the collected data, and evaluating different policies in the real world. For the third step, we followed the same approach as in \cite{khansari2022practical}. For the models in the policy testing phase, matching colors indicate that they were tested on the same set of real world doors.}
\vspace{-5mm}
\label{workflow}
\end{figure*}

Given a collected dataset, a secondary, yet critical, challenge is how to remove low-quality data or mislabeled data from the dataset. Prior work curated data manually by manual examination of the dataset \cite{friedrich1995robot}. Researchers have used statistical methods such as Low Confidence Correction and Jitter \cite{chuck2017statistical} and ActiveClean \cite{krishnan2016activeclean} to filter noisy data. The data curation approach in this paper used a combination of manual and statistical curation to improve performance. Furthermore, researchers explored haptic feedback during different phases of LfD. In prior work, researchers aimed to learn robot manipulation policies only through forces \cite{rozo2013robot}. They used a high-fidelity kinesthetic haptic device during example collection, in combination with other novel data encoding techniques, and found the robot could reach comparable human teacher performance on two manipulation tasks. Other researchers equipped a robot with a pressure cuff so human demonstrators could encode stiffness profiles during example collection; they concluded that adding the stiffness profile to the control scheme increased the robot's success rate at lighting a match \cite{kronander2013learning}. In different work, researchers used kinesthetic teaching to demonstrate a position profile and then combined this with teleoperation data collected via a kinesthetic haptic device to teach a force profile \cite{kormushev2011imitation}. Researchers trained a robot to provide real-time haptic guidance to a surgical demonstrator during cooperative control between a human and Raven II robot \cite{power2015cooperative}. Prior work thus indicated that kinesthetic haptic feedback during example collection can improve robot performance, but has not thoroughly examined if and how haptic feedback is useful when training a deep imitation learning model, training an image-only model (versus one with force information), or collecting data with low-cost, ungrounded haptic devices. \newchange{Li et al.~\cite{li2023immersive} addressed this by using a real-world haptic device to collect data to train a deep imitation learning model, with their data collection and policy execution performed in simulation on a pick-and-place task. In the current paper, we leverage haptic feedback during the data collection phase with a real-world robot to improve the demonstration data throughput and its quality for the purpose of training a deep visual imitation learning model.}


\section{Methods}
\label{sec:Methods}

In order to address our primary research questions, the project was divided into two phases. Phase 1 studied demonstrators' performance with and without haptic feedback and Phase 2 analyzed the real-world performance of the trained policies on opening latched doors autonomously. Figure \ref{push_door_setup} shows a demonstrator opening the door via teleoperation and zoomed-in images of the gripper movement. In this task, the demonstrator stood behind the robot while it moved to a starting configuration, and then teleoperated the arm to open the door. After this, the demonstrator marked the task success or failure. Figure \ref{workflow} shows the variables and their range of values for our study. The two feedback conditions for teleoperation were \change{Haptic and Non-Haptic}. \change{The study began with nine human demonstrators, though three were excluded early in the experimental process}. Data was collected on four doors and policies were tested on all four doors seen during training and an additional two doors for each side (left and right). All doors were in the same building.

\change{We used a 9-degree-of-freedom (DoF) prototype mobile manipulator. This robot was equipped with a two-finger gripper, a custom 6-axis force-torque sensor in the wrist, and a camera in the head (referred to as the first-person camera). The  2-DoF robot base was fixed during data collection, but it was used for teleoperating the robot to a location close to the door during the experiment (before data collection began). The robot was teleoperated using Oculus Quest controllers from the commercially available Oculus Quest virtual reality headset system.} There were two controllers, one for each hand of the demonstrator; the left-hand controller moved the robot's base and the right-hand controller moved the robot's arm. Teleoperation control was done by mapping the controller 3D pose to the robot's gripper in Cartesian space and then performing inverse kinematics to get the robot joint angles. In all conditions, the human demonstrators had a direct line of sight with the robot and the door. The teleoperation controller had a single linear vibration actuator in each hand controller.

\newchange{When provided, haptic feedback in the form of 200 Hz vibrations was provided by the controller held in the right hand. Larger magnitude forces were mapped to higher-amplitude vibrations. We measured the magnitude of the contact force using the force-torque sensor in the robot’s wrist. Vibrations are rendered on the Oculus Quest controller by mapping an input value of 0-255 proportionally to a vibration output range of 0-1.76 g in amplitude. Thus, we multiplied the measured force magnitude by a constant gain to give an intermediate number between 0-255 for input to the Oculus Quest controller. When the force magnitude was less than 2N, the Oculus Quest output was not perceivable. When the force magnitude was over 24N, we sent the maximum amplitude. The haptic feedback experience was tested and tuned by the researchers and a demonstrator who did not participate in the study prior to being deployed on the controller. The amplitude algorithm is shown in Algorithm \ref{amplitude_algorithm}.}

\begin{algorithm}
    \caption{Haptic amplitude algorithm}\label{amplitude_algorithm}
    \begin{algorithmic}
        \State  $forceMagnitude = \sqrt{x^2 + y^2 + z^2}$
        \State $gain = 9$
        \If{$forceMagnitude < 2N$} 
            \State amplitude $= 0$
        \ElsIf{$forceMagnitude >= 24N$}
            \State amplitude $= 1.76 g$
        \Else 
            \State amplitude $= (forceMagnitude*gain*1.76)/255$
        \EndIf

    \end{algorithmic}
\end{algorithm}

\newchange{For the autonomous policy execution, the closed control loop runs at a rate of 10 Hz. Prior work from this team (Figure 9 in \cite{khansari2022practical}) describes the robot control system and network architecture in detail. In summary, we pass  RGB images from the head-mounted camera to a ResNet-18 encoder that projects the mean pool layer to three ``action heads'' (predicted actions).} 

\subsection{Phase 1: Studying haptic feedback during demonstrator performance}

Two datasets were collected for opening a door by pressing down on the latch-type handle and pushing the door open: with haptic feedback and without haptic feedback. \newchange{These datasets were collected over two days for each participant -- one day per condition.} The nine demonstrators were professional robot teleoperators who had used the robot previously for various tasks. All demonstrators were right-handed and their experience varied from 2 months to 18 months worth of experience in this role. Regardless of their experience, each demonstrator received one hour of practice on the task prior to commencing formal data collection. \change{This practice hour was necessary in order for the demonstrators to learn how to launch this modified teleoperation application, record their work, and learn the button mappings on the teleoperation controller.} All demonstrators completed a short interview at the start of the first day in order to describe their past experience with this robot and their past teleoperation experience. Nine demonstrators participated in the task. There were five men, one woman, one non-binary person, and two people declined to state their gender. Seven of the participants were 25-34 years old, one person was 35-44 years old, and one person declined to state. \change{The Stanford University Institutional Review Board (IRB) determined that this study does not meet the regulatory definition of human subjects research; it was exempt from IRB approval because the participants were paid demonstrators performing this teleoperation work as their job. The inclusion criterion was that the participants had to be professional demonstrators with experience teleoperating the type of robot used in the study.}

\subsubsection{Real-world data collection}

Teleoperation data was collected for the task of opening push latch doors. The task was as follows: the demonstrator stood behind the robot while the robot's arm moved to a starting configuration near the front of the door. \change{The demonstrators were in the same physical space as the robot and thus had a direct visual of the robot and environment. The demonstrators are not constrained to only the robot camera view since it lacks depth information.} Recall that the controller position was mapped to the position of the robot's gripper. The demonstrator began recording an episode, and moved the controller through space so the robot's gripper reached the door handle. The demonstrator then pushed down and released the latch. The demonstrator attempted this until they reached either a success (door unlatched) or failure state, and then marked the episode immediately after completing it. \change{The demonstrator could make as many movements as they desired during a single example collection.} Demonstrators marked the episode as a failure if the fingers fell off, the robot entered an error state, the arm was in a strange position that was hard to control, or the robot remained in prolonged contact with the latch after the door opened. \change{An example is marked successful when the door was open more than $1$ degree and the robot was no longer in contact with the door}. The base of the robot remained stationary for the duration of the episode collection. The demonstrators were instructed to collect $\sim 200$ successful examples or stop at the end of an eight hour workday if they did not collect this number. They spent the full day of data collection in the same condition: either \change{Haptic or Non-Haptic}. Conditions were randomized so demonstrators varied as to which condition they received first and second.

The summarized participant task instructions are below:

\textit{Today you will be collecting data on push latch door opening. Your goal will be to open the door as gently and efficiently as possible. For the latched door, we do as follows:
Press the “clutch” or middle finger button on the right controller to move the robot. While holding the clutch, place one finger on top of the latch and press straight down. If the door is unlatched, i.e. can be pushed open without turning the handle again, it is a success. If the fingers snap off, or the robot enters an error state, or the robot stays in contact with the door after opening it, it is a failure. If you are using the haptics application you will feel some buzzing in your hand when you come into contact with objects.
When you feel this buzzing become MUCH stronger, it means you are surpassing the safe force limits of the robot, back away and try again.}

Data was collected on four meeting room doors in an office building. All demonstrators aimed for an even distribution of right swing and left swing doors, meaning half the total data collected was left swing and half the total collected was right swing. Some demonstrators were less successful with opening left swing doors than right swing doors; therefore their collected number of examples may differ for right swing vs. left swing doors.

The base was placed approximately 0.5 meters away from the door. There was variation in the placement of the base relative to the door due to the inherent imprecision in navigating the robot in front of the doors. In addition, the base occasionally moved forward unintentionally due to the force of the robot's gripper on the door. The arm was moved to a predefined initial joint configuration at the start of the episode, one configuration for left swing doors and one configuration for right swing doors. Left swing doors are doors that have hinges on the left (when facing the door) and right swing doors are doors with hinges on the right. RGB images from the head camera were recorded for input to the model. \newchange{Haptic or force information were \emph{not} fed to the model. The action space was 7-dimensional, comprised of the arm motion (7-DoF) in Cartesian coordinates during data collection; for the policy run, the action space is also 7-DoF, but in joint space.} 

At the end of each day, each demonstrator filled out a survey with the NASA task load index (a standardized set of questions asking about the perceived difficulty of a task \cite{hart2006nasa}) and qualitative feedback questions. The survey consisted of the following 6 questions and the response to each question was on a scale of 0-10, where 0 was very low and 10 was very high. The questions were: (1) How mentally demanding was the task?, (2) How physically demanding was the task?, (3) How hurried or rushed was the pace of the task?, (4) How successful were you in accomplishing what you were asked to do?, (5) How hard did you have to work to accomplish your level of performance?, and (6) How insecure, discouraged, irritated, stressed, and annoyed were you?. The demonstrators also completed qualitative questions like ``Describe the strategy you used to open the door'' and ``Do you have any comments to share about the task?''.

\subsubsection{Data curation and model training}

\begin{figure*}
    \centering
    \includegraphics[width=\linewidth]{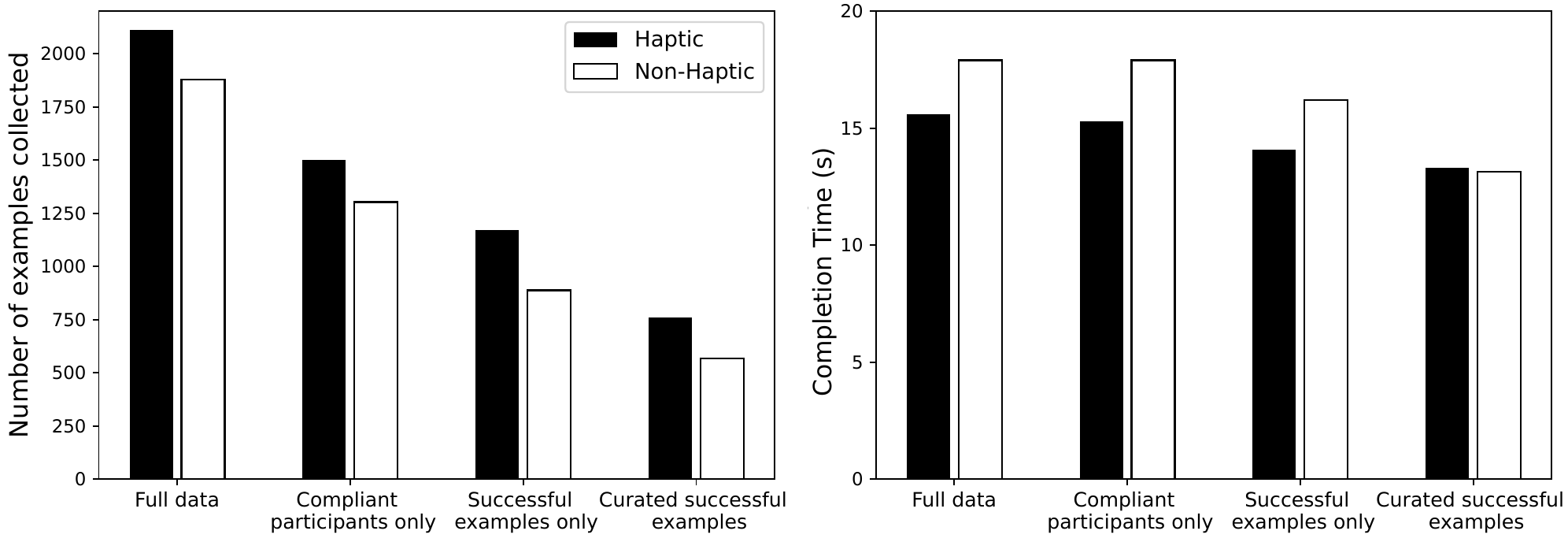}
    \caption{\change{Nine demonstrators represented 100\% of the original data in the ``Full data'' bars. Three demonstrators were removed due to non-compliance in the ``Compliant participants only'' bars. Failures were removed from remaining six demonstrators' data to create the ``Successful examples only'' subset. The remaining data was further curated by duration cutoff, and undesirable behavior removal resulting in the final ``Curated successful examples''. At left, a larger percentage of ``Curated successful examples'' remained in the \change{Haptic} condition, showing the improvement in data quantity. At right, demonstrators took less time to complete the task when receiving haptic feedback in all circumstances except when narrowing the data to ``Curated successful examples''.}}
    \vspace{-5mm}
    \label{examples_collected}
\end{figure*}

Demonstrators varied in the amount of successful data they collected in a full day of work due to their experience level. In Figure \ref{workflow}, the first removal criterion was failure to follow instructions (``Compliant participants only''). Under this criterion, three demonstrators repeatedly failed to follow instructions. These three demonstrators were excluded from all datasets for model training. \change{If these demonstrators had been compliant, their data would have then been cleaned with the rest of the data (removing failures next). As is standard practice in LfD, we removed the failures from the dataset. The second removal criterion was eliminating failures (``Successful examples only'').} Recall that the demonstrator marked the episode a failure if the robot stayed in contact with the door after opening it, the fingers snapped off of the wrist, or the robot entered an error state \newchange{(either due to demonstrator error or technical problems)}. After removing all the failures from the dataset, the \change{trial} models were trained only with success data from the six remaining demonstrators. \change{These trial models were not used for real-world policy testing.}

After this \change{trial} model training, more data curation was necessary due to the wide variance in demonstrator behavior and approaches to the task. \newchange{Thus, the third removal criterion (``Curated successful examples'') was based on the following: removing episodes longer than 20 seconds (which is approximately 50\% longer than the time to complete a successful episode), removing mislabeled episodes including those where the robot gripper collided with the door, removing episodes where the demonstrator pulled up on the door handle rather than pushing down, and removing episodes where the demonstrator failed to follow the task instructions.} \newchange{The Completion Time was less after this third removal criterion was applied because there were a larger number of long episodes for ``Successful examples only''.}

\subsection{Phase 2: Robot performance evaluation}

In the second phase, we examined the effect of data quality on autonomous robot performance in the real world. Identical model architectures and evaluation methods were deployed with each dataset. 

\subsubsection{Model description and checkpoint selection}

The aim throughout model training was to learn a policy  $\pi(a\vert s)$ that output a continuous action $\textit{a} \in \textit{A}$ given an RBG image describing the state $\textit{s} \in \textit{S}$. A set of demonstrations was described as $\tau^* = (s_0, a_0, s_1, a_1...s_{T-1}, a_{t-1}, s_T)$ and the learned policy was $\pi^*$. With the set of demonstrations, we minimized $\pi(a|s)-\pi^*(a|s)$ given the same state $s$. We followed the approach taken in \cite{khansari2022practical} and used a visual deep imitation learning model to train the policies. In this architecture, an RGB image was fed as input to a ResNet-18 model followed by a mean-pool and unit norm layers to create the visual embeddings. They were then fed to two action heads: the arm joint angles (7-DoF), and terminate action (1-DoF). The data preprocessing and augmentation included random cropping of RGB images and photometric distortion (similar to \cite{young2021visual, khansari2022practical}).

Six models were trained with the curated data: haptic feedback and left swing doors only, haptic feedback and right swing doors only, haptic feedback and joint right/left swing doors, no haptic feedback and left swing doors only, no haptic feedback and right swing doors only, no haptic feedback and both right/left swing doors, see the Model Training and Evaluation section of Figure \ref{workflow}. As we will discuss in the next section, the \change{Haptic} dataset was larger than the \change{No Haptic} dataset. To better analyze the effect of data quality on the model performance, all the \change{Haptic} datasets were down-sampled to be at the same size as the corresponding \change{Non-Haptic} datasets, \change{both datasets were comprised of the same demonstrators}. The models trained for 20,000 steps and 34 checkpoints were saved and evaluated in simulation. A checkpoint was comprised of the exported trained weights of the model at a given training step. Recall that we only trained models on the real world collected data. We used the \change{real-to-sim} head of a pretrained RetinaGAN model, as in \cite{ho2021retinagan}, to adapt real images to look like simulated images in order to enable evaluations in simulation. The starting $X, Y$ simulation position of the base and the yaw angle $\theta$ was randomized over a normal distribution that matches the real world robot initial pose configuration. An episode success during checkpoint evaluation was determined with the same criteria used during data collection: the robot should not be in contact with the door, and the door should be open more than $1$ degree.

\subsubsection{{Policy testing on real doors}}

\begin{figure}[t]
\centering
\includegraphics[width=\columnwidth]{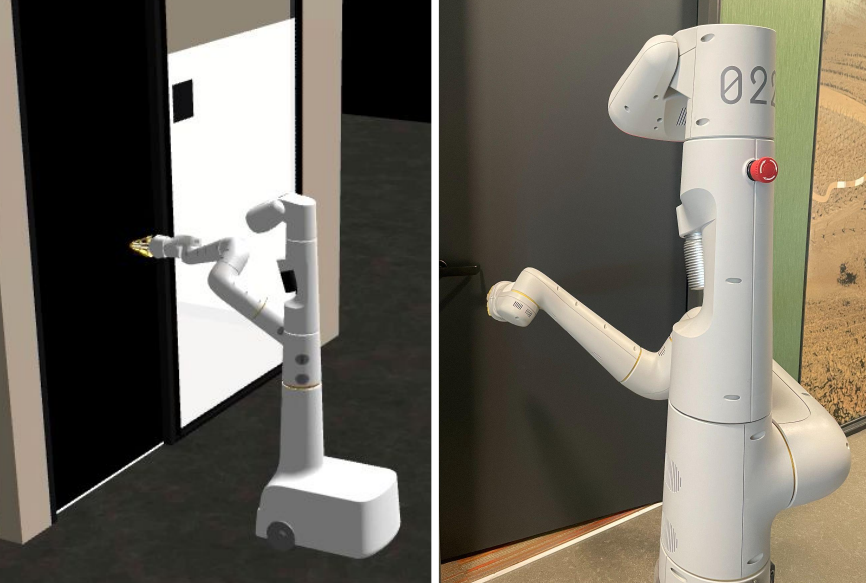}
\caption{A checkpoint was comprised of the exported trained weights of the model at a training step. Checkpoints were first evaluated in simulation (left), and then the top three best checkpoints were used for real world model evaluation (right).}
\vspace{-5mm}
\label{simscene}
\end{figure}

The three highest-performing checkpoints in simulation were selected for policy testing in the real world. The selected checkpoints were tested on real robots at the same building where the original data was collected. Figure \ref{simscene} shows the robot in simulation and real world, opening a door autonomously in both cases. The same type of robot that was used in data collection was used for policy evaluation. The policies were tested for 20 episodes on each door at a two-inch distance. The number of seen (included in the training data) and unseen doors tested for each model are summarized in Figure \ref{workflow}. The placement of the robot base was randomized the same way in the real world as it was during simulation evaluation. There was no randomization in the arm starting position, matching our data collection protocol. For each tested autonomous episode, the episode was automatically marked a failure after one minute. Similar to data collection, the episode was also marked a failure if the robot stayed in contact with the door after opening it, if the fingers snapped off of the wrist, or if the robot entered an error state. \change{The best performing checkpoints in simulation for each model type had the following \% successful trials: Haptic Left 87.5\%, Haptic Right 60.0\%, Haptic Left \& Right 68.1\%, Non-Haptic Left 91.2\%, Non-Haptic Right 58.8\%, and Non-Haptic Left \& Right 50.6\%.}


\vspace{-3mm}
\section{Results}
\label{sec:result}

\subsection{Phase 1: Impact of haptic feedback on demonstrator performance}

The \change{Haptic} datasets began with more data, as shown in Figure \ref{examples_collected}, and a larger percentage of this data remained after curation. The \change{Haptic} condition resulted in a higher percentage of curated data as a proportion of total data collected for 9 demonstrators: 36\% of \change{Haptic} data remained after curation versus 30\% for the \change{Non-Haptic} dataset. For the final curated data as a proportion of the 6 demonstrators, 51\% of \change{Haptic} data remained after statistical curation, while 43\% of \change{Non-Haptic} data remained. Demonstrators with haptic feedback took less time to perform the task in all data subsets except ``Curated successful examples'' where the durations were similar.

\change{The overall results from Phase 1 indicated that demonstrators performed better and preferred teleoperation with haptic feedback. Demonstrators in the Haptic condition had a 9.5\% higher success rate when performing the push door task than those without feedback, as shown in Figure \ref{successandtime}. We \change{used a linear mixed model \newchange{and Likelihood Ratio test} to study the effect of Haptic/Non-Haptic conditions on the success rate with and it was statistically significant ($p<0.001$, F = 38.07)}. This 9.5\% improvement is greater than the 3\% task experience improvement from all day 1 data (``1'') to all day 2 data (``2'') which was also not statistically significant. Comparing \change{Haptic and Non-Haptic} conditions with the first and second day of the task, the individuals with haptic feedback performed the best across all four subsections. Individuals with haptic feedback took slightly less time than those without haptic feedback in order to successfully complete the task, shown at right in Figure~\ref{successandtime}. This may have been because the demonstrator better knew when they are contacting the door and could have more quickly removed the robot's gripper from contact once the door had unlatched. We also found a statistically significant effect of Left or Right swing doors on success ($p < 0.001$, $F = 21.48$).}

\begin{figure}[t]
\centering
\includegraphics[width=\columnwidth]{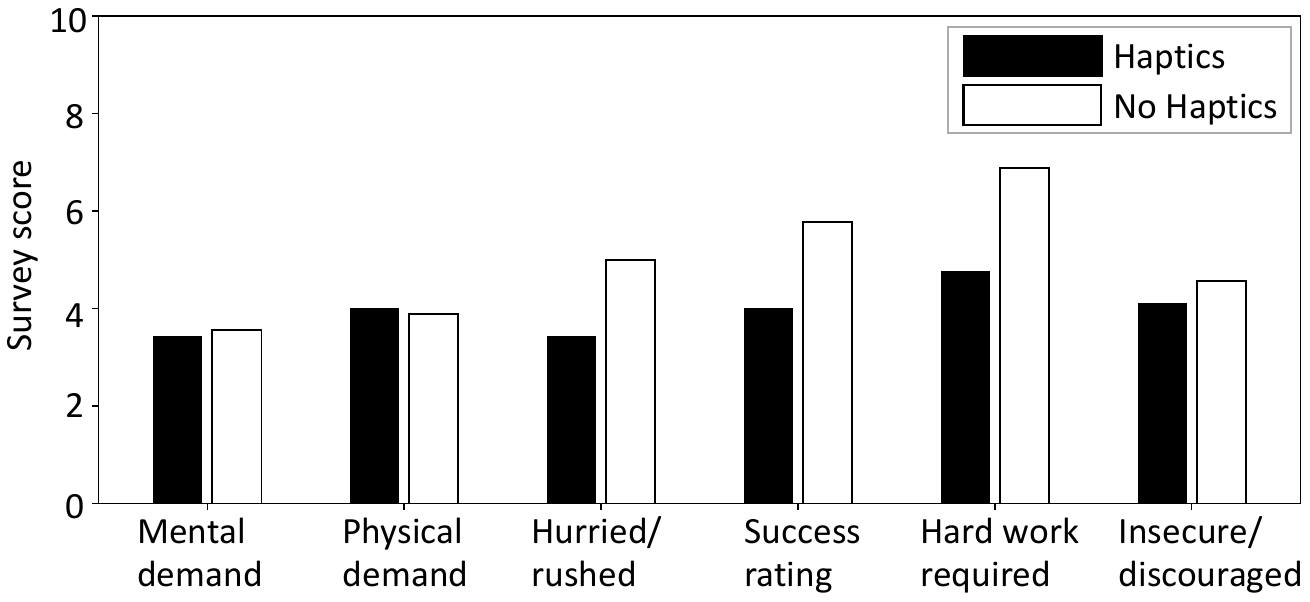}
\caption{\newchange{Results of the NASA TLX survey conducted in Phase 1 (data collection).}}
\vspace{-5mm}
\label{surveyscore}
\end{figure}

Demonstrators completed the NASA TLX survey at the end of the day of data collection. They rated the data collection without haptics to be more mentally demanding, hurried, annoying/irritating, and requiring more work, and considered themselves less successful at the task. The only category in which haptics was rated less favorably was that it required more physical demand. \change{A within-subjects ANOVA was performed across these categories to study the effect of Haptic or Non-Haptic condition on question responses. Question 6 (insecure) was significant at $p < 0.1$, Question 5 (hard work) was approaching significance, and the other differences were not statistically significant.} \newchange{We posit that the demonstrators experienced the Haptics condition as more physically demanding (although it was not a statistically significant difference), because there was more sensoral engagement (both vision and haptics rather than only vision) in the Haptics condition.} The difference in reported difficulty is supported by the collected data: demonstrators were able to collect more data when receiving haptic feedback, see left side of Figure \ref{examples_collected}. For 9 demonstrators, this is statistically significant with $p<0.1$. Demonstrators 7-9 were eventually removed in the manual curation, leading to the same conclusion with $p<0.05$. Further analysis of these demonstrator reports are in the discussion. These data demonstrate that example collection with haptic feedback leads to more data throughput, as more data is collected and more curated data remains. Providing the haptic feedback to the demonstrator changes their behavior by providing a new sensory modality in addition to vision. This improves the quality of the data that they collect. 

\begin{figure*}
\centering
\includegraphics[width=\linewidth]{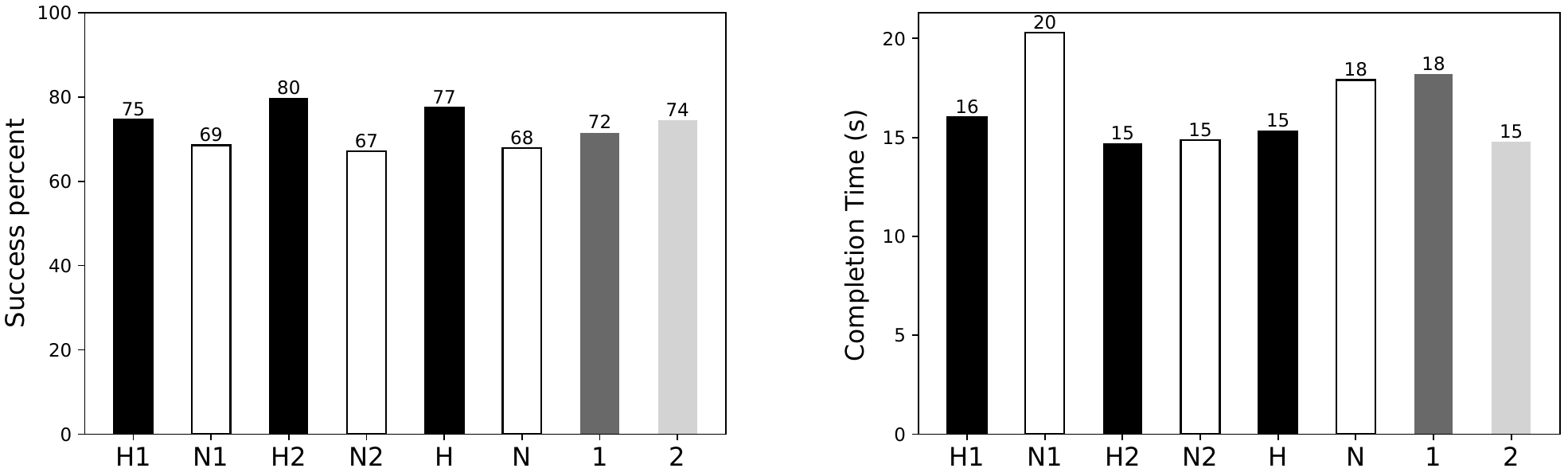}
\caption{\change{\newchange{Demonstrator performance during Phase 1 (data collection).} At left, demonstrators were overall 9.5\% more successful when receiving haptic feedback (``H'') than no haptic feedback (``N''). Both day 1 and day 2 of haptic feedback (``H1'', ``H2'') showed higher success rates than day 1 and day 2 of no haptic feedback (``N1'', ``N2''). At right, haptic feedback allowed the demonstrator to complete the task in 4 seconds less time on the first day (``H1'') than without haptic feedback (``N1''). However, the difference in time needed to complete the task was greater from day 1 (``1'') to day 2 (``2'') than from \change{Non-Haptic} (``N'') to \change{Haptic} (``H'').}}
\vspace{-5mm}
\label{successandtime}
\end{figure*}

Comments from demonstrators who preferred haptics included, ``Buzzing is really helpful when ever robots gripper touches the door handle and the doors'', ``I prefer the haptic feedback'', ``The haptic helps knowing how much pressure you are applying. Very hard task as the doors are hard to unlatch'', and ``The task was a little harder without the feedback, having it really gave me a sense of contact compared to my own sight with controllers.'' Other comments from demonstrators who did not prefer haptics included, ``I did not notice too big of a difference with haptic feedback vs non-haptic feedback in terms of success rate'' and ``I don't think the haptic feedback made much of a difference but it was nice to know when there was a lot of pressure on the arm.''
\vspace{-3mm}

\begin{figure*}
\centering
\includegraphics[width=\linewidth]{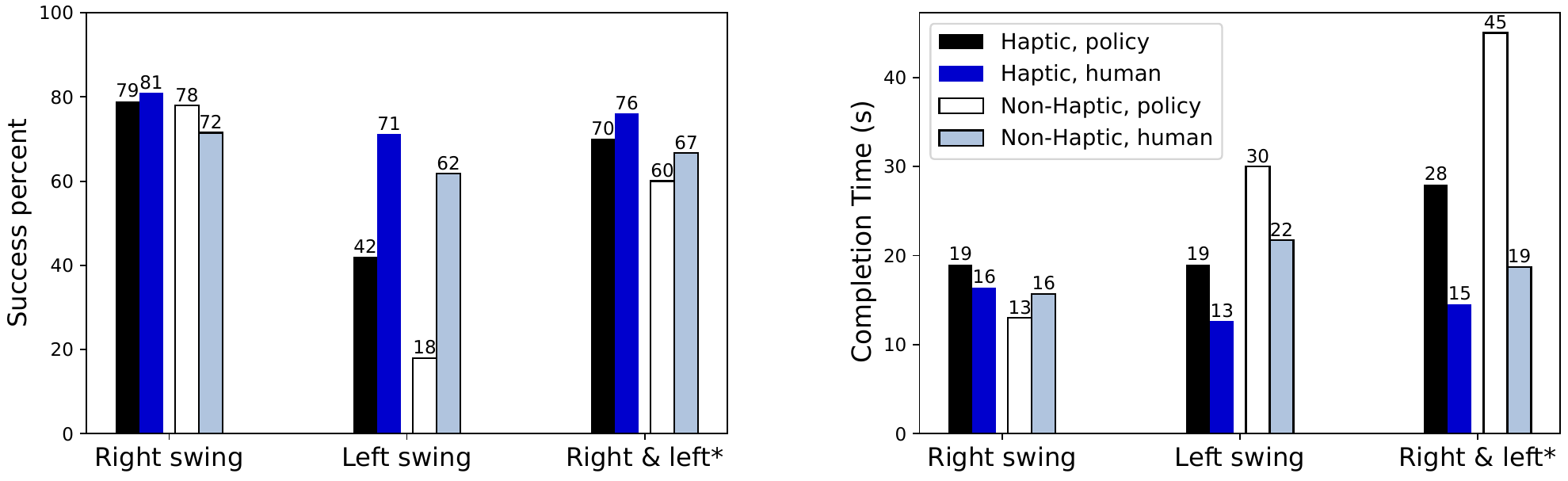}
\caption{\change{\newchange{Autonomous robot performance during Phase 2 (robot performance evaluation).} Evaluations of policy performance on real doors for six trained models alongside human demonstrator performance. As shown on the far left, the \change{Haptic and Non-Haptic} policies for Right swing doors performed similarly. However, the \change{Haptic} policies for Left swing doors and the combined Right \& left swing doors performed better than the \change{Non-Haptic} policies for these cases. Thus, overall, policies trained with \change{Haptic} data performed better than policies trained with \change{Non-Haptic} data. \textit{(* indicates that the human data in this category were an average of Right swing and Left swing door data, whereas  the policies were trained with a combination of Right and Left swing door data.)}}}
\vspace{-5mm}
\label{allevals}
\end{figure*}

\vspace{-2mm}
\subsection{Phase 2: Impact of \change{Haptic and Non-Haptic} datasets on autonomous robot performance}

All six policies were deployed in the real world and tested at four doors relevant to that side: two unseen doors and two seen doors. For example, a policy trained on only right swing doors would be tested on two unseen right swing doors and two seen right swing doors. The unseen doors were included to test the generalization capabilities of the policy. The policies trained on both left and right swing door data were tested on eight doors -- four seen doors and four unseen doors. For right swing doors only (at the left in Figure \ref{allevals}), policies trained both with and without haptic feedback performed well in the real world, at 79\% and 78\% success, respectively. The demonstrators noted that this task was easier than the left swing door task, thus we can infer that the haptic feedback did not have much of an effect. For models trained on the left swing doors the policy trained with the \change{Haptic} dataset performed 24\% better than the policy trained with the \change{Non-Haptic} dataset. For each door side and each policy condition, policies performed slightly better on the seen doors rather than the unseen doors. For the two policies trained on a combination of left and right swing doors in Figure \ref{allevals}, the policy trained with haptic feedback examples performed 10\% better than the policy without haptic feedback. This was consistent with the differences between the left swing and right swing door side-specific policies. We observed a higher average policy performance when combining the left and right data, meaning there was some transfer learning between the left and right.

\change{The data were analyzed with a chi-square test of independence performed on all Haptic and Non-Haptic policy trials, regardless of the door side. There was a statistically significant difference between Haptic and Non-Haptic policies with $p < 0.05$. Separate, individual chi-square tests of independence were performed comparing each door condition (Left, Right, or Right \& left) and Haptic or Non-Haptic policies. The Left Haptic policy was statistically significantly different than the Left Non-Haptic policy with $p < 0.05$. This test for independence was statistically significantly different with $p < 0.1$ for the Haptic vs. Non-Haptic Right \& left door combination and was not statistically significant for the Haptic vs. Non-Haptic Right swing doors. This shows that haptic feedback not only leads to better demonstrator performance for more challenging tasks, but also leads to improved policy performance for more challenging tasks.} 

\change{The demonstrators were all right-handed, which may account for the difference in success rate between the right and left swing doors. As shown in Figure \ref{allevals}, regardless of training dataset, the policy performance was similar to the human demonstrator performance for right swing doors, but the human demonstrator performance exceeded the policy for left swing doors.} Comparing all training data conditions -- right only, left only, and right and left -- the policy trained with haptic feedback led to an overall 11\% improvement in performance. Thus, haptic feedback led to higher data quality and a better-performing autonomous robot policy. 
\vspace{-3mm}


\section{Discussion and Conclusion}
\label{sec:conclusion}

Our results indicate that haptic feedback can be useful for robot imitation learning. We showed this utility at two points in the imitation learning process: (1) Data collection and (2) Policy performance, \newchange{we also performed both of these phases in the real world}. In (1), haptic feedback improved demonstrator performance during data collection by increasing the success rate, meaning fewer hours were needed overall to collect a given number of successes. \change{We showed this effect with a low-cost, readily-accessible teleoperation controller.} In addition, haptic feedback improved the amount of high-quality data gathered during data collection, as there was more proportional data remaining from the haptic feedback dataset after the curation. \change{Through NASA TLX qualitative ratings, the demonstrators in this study preferred haptic feedback, a finding that could impact cases where demonstrator fatigue is costly.} Demonstrator performance and preference continues to be important while increasing the amount of tasks or needing more high quality data to train a viable policy.

In (2), the improved data manifested in policy performance improvements. After applying the data curation as described in Section 3.1.2, the performance difference between the two conditions \change{(Haptic and Non-Haptic)} still remained. This suggests that both the haptic feedback and data curation contributed to better performance. When comparing autonomous model performance in the real world with the same amount of curated data, policies trained with haptic feedback examples performed far better on left swing door-specific policies and on policies jointly trained with left and right swing door data. The \change{Haptic and Non-Haptic} policies trained on right-only data were similar. One possible explanation for these results is the left-sided door task was rated more complex by the demonstrators, thus they benefited more from the additional information channel of haptic feedback. \newchange{In cases where demonstrators are already very proficient at a task, they may not benefit as much from the additional feedback.} \change{In addition, the models trained for policy testing did not include force information as inputs, showing that changing the demonstrator behavior without altering the model architecture can result in significant changes to autonomous robot performance.} Thus, we conclude that adding haptic feedback to teleoperation can be meaningful in a robot learning setting where ample quantities of high quality data are necessary.

\bibliography{references}

\bibliographystyle{IEEEtran}

\begin{IEEEbiography}[{\includegraphics[width=1in,height=1.25in,clip,keepaspectratio]{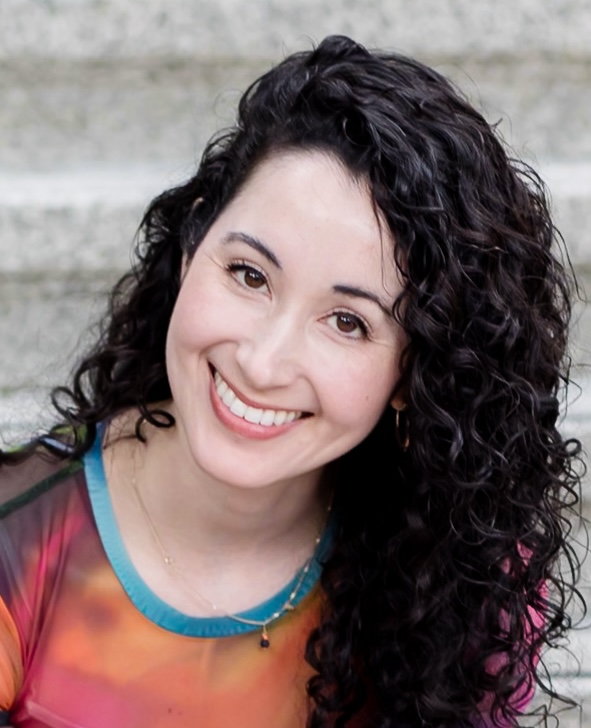}}]{Catie Cuan} received M.S. and Ph.D. degrees from Stanford University in mechanical engineering, where she is currently a Postdoctoral Researcher in computer science. She is also the CEO of an early-stage, venture backed AI company. Her research interests include haptics, imitation learning, human-robot interaction, performance, and art.
\end{IEEEbiography}

\vspace{-4mm}

\begin{IEEEbiography}[{\includegraphics[width=1in,height=1.25in,clip,keepaspectratio]{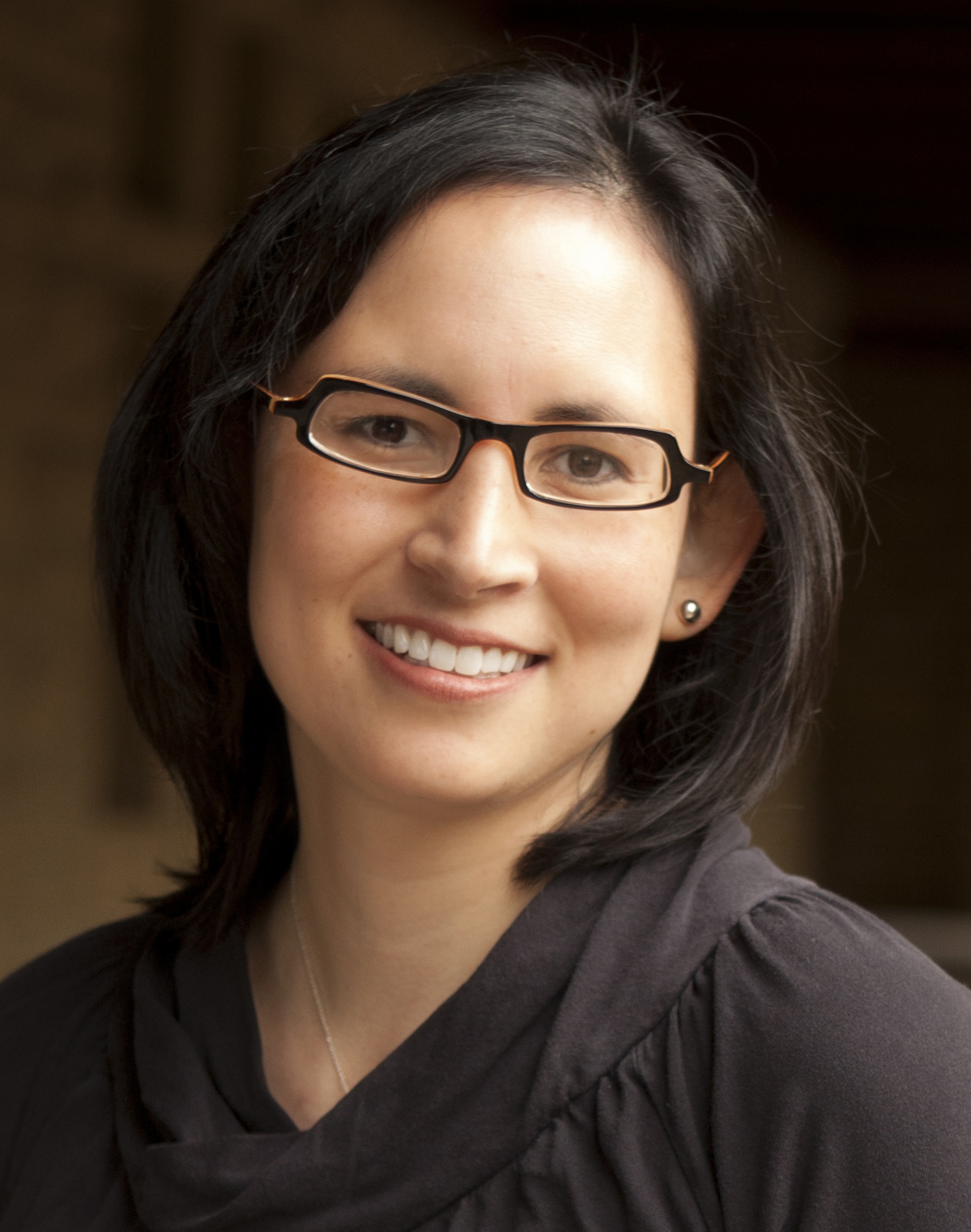}}]{Allison M. Okamura} (Fellow, IEEE) received the B.S. degree from the University of California Berkeley in 1994, and the M.S. and Ph.D. degrees from Stanford University, Stanford CA, in 1996 and 2000, respectively, all in mechanical engineering. She is currently the Richard W. Weiland Professor in the School of Engineering and Professor of Mechanical Engineering at Stanford University, Stanford, CA. Her research interests include haptics, teleoperation, medical robotics, virtual environments and simulation, neuromechanics and rehabilitation, prosthetics and engineering education.
\end{IEEEbiography}

\vspace{-4mm}
\begin{IEEEbiography}[{\includegraphics[width=1in,height=1.25in,clip,keepaspectratio]{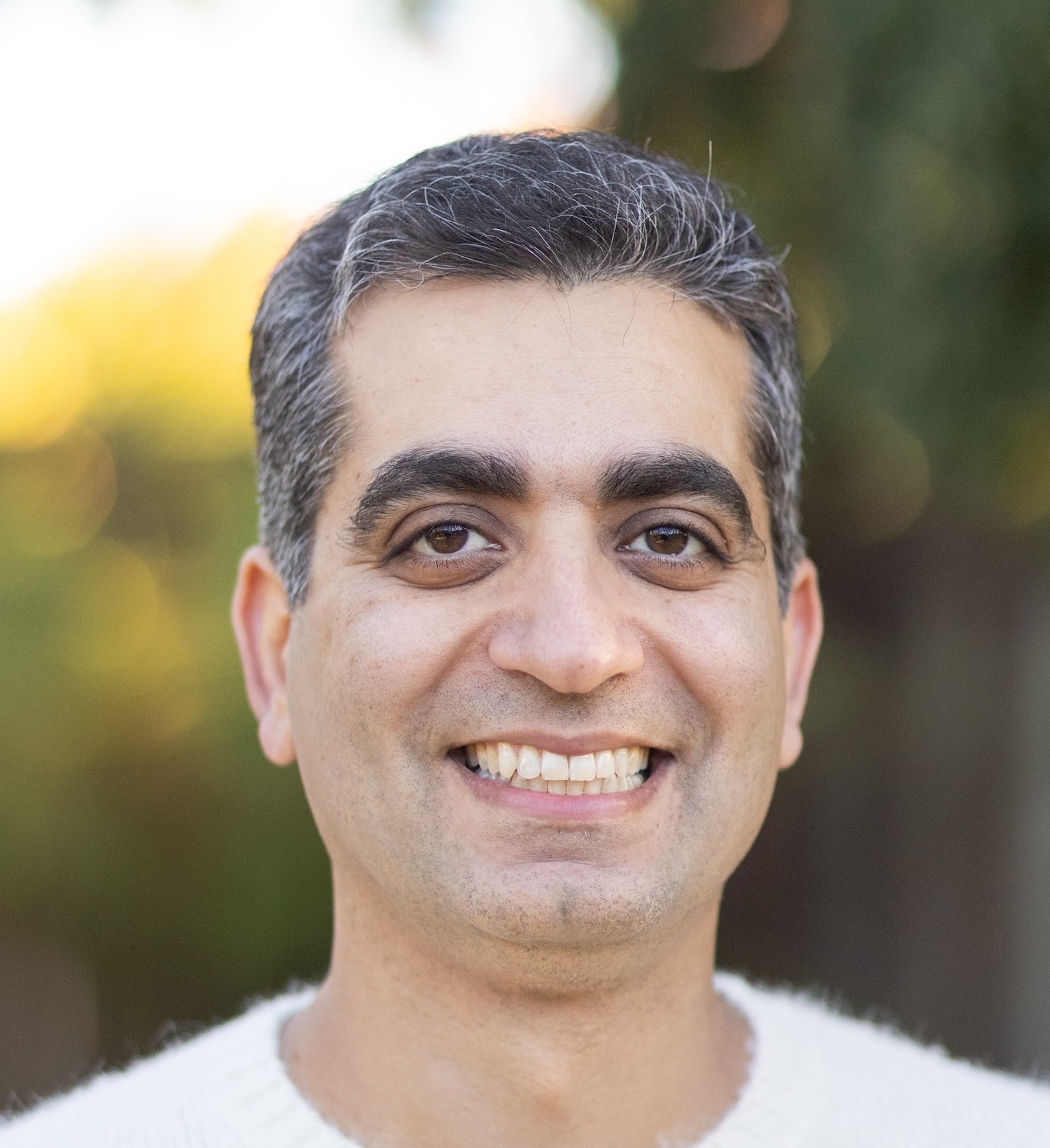}}]{Mohi Khansari} is currently a Sr. Staff Applied Scientist at Cruise. Prior to that, he was a Tech Lead and Staff Roboticist at Everyday Robots, a Google X moonshot project. Mohi received his PhD from Ecole Polytechnique Federale de Lausanne (EPFL) in 2012 in the field of Manufacturing Systems and Robotics. He was also a Postdoc researcher at Stanford AI Lab from 2014 to 2016. Mohi's interests include Deep Visual Imitation Learning, Learning from Demonstrations, Dynamical Systems, and Control.
\end{IEEEbiography}

\end{document}